\documentclass[a4paper]{article}

\usepackage[full]{textcomp}
\usepackage{INTERSPEECH2019}
\usepackage{amsmath,graphicx}
\usepackage{epsfig}
\usepackage{xspace}
\usepackage{srcltx}
\usepackage{caption}
\usepackage{subcaption}
\usepackage[export]{adjustbox}
\usepackage{multirow,makecell,hhline}
\usepackage{booktabs}
\usepackage{cite}
\usepackage{hyperref}
\usepackage{url}
\usepackage{CJKutf8}
\usepackage{paralist}
\usepackage{newtxtext,newtxmath}

\usepackage{color}
\usepackage[dvipsnames]{xcolor}

\newcommand{\CMT}[1]{{}}

\newcommand{\tbh}[1]{#1}

\def\L{{\cal L}}

\def\L{{\cal L}}

\newcommand{\dg}[1]{{\color{darkgray}#1}}

\newcommand{\smallurl}[1]{\scriptsize \url{#1}}
\newcommand{\cmmnt}[1]{}
\title{Learning to Speak Fluently in a Foreign Language:\\
Multilingual Speech Synthesis and Cross-Language Voice Cloning}

\name{Yu Zhang, Ron J. Weiss, Heiga Zen, Yonghui Wu,  Zhifeng Chen, RJ Skerry-Ryan, Ye Jia, \\ Andrew Rosenberg, Bhuvana Ramabhadran}
\address{Google}
\email{\{ngyuzh, ronw\}@google.com}

\begin{document}
\ninept
\maketitle
\begin{abstract}
We present a multispeaker, multilingual text-to-speech (TTS) synthesis model based on Tacotron that is able to produce high quality speech in multiple languages.
Moreover, the model is  able to transfer voices across languages, e.g.\ synthesize fluent Spanish speech using an English speaker's voice, without training on any bilingual or parallel examples.
Such transfer %
works across distantly related languages, e.g.\ English and Mandarin.

Critical to achieving this result are: \begin{inparaenum}[1.]\item using a phonemic input representation %
to encourage sharing of model capacity across languages,
and %
\item incorporating an adversarial loss term to encourage the model to disentangle its representation of speaker identity (which is perfectly correlated with language in the training data) from the speech content.
\end{inparaenum}
Further scaling up the model by training on multiple speakers of each language, and incorporating an autoencoding input to help stabilize attention during training, results in a model which
can be used to consistently synthesize intelligible speech for training speakers in
all languages seen during training, and in native or foreign accents.

\end{abstract}

\section{Introduction \label{sec:introduction}}

Recent end-to-end neural TTS models~\cite{van2016wavenet, wang2017tacotron, arik2017deep} have been extended to enable control of speaker identity \cite{arik2018neural,jia2018transfer,nachmani2018fitting,chen2018sample} as well as unlabelled speech attributes, e.g.\ prosody, by conditioning synthesis on latent %
representations~\cite{wang2018style, skerry2018towards, akuzawa2018expressive, henter2018deep, hsu2018hierarchical} in addition to text.
Extending such models to support multiple, unrelated languages is nontrivial %
when using language-dependent input representations or model components,
especially when the amount of training data per language is imbalanced.
For example, there is no overlap in the text representation between languages like Mandarin and English.
Furthermore, recordings from bilingual speakers are expensive to collect.  It is therefore most common for each speaker in the training set to speak only one language, so speaker identity is perfectly correlated with language.
This makes it difficult to transfer voices across different languages, a desirable feature when the number of available training voices for a particular language is small.
Moreover, for languages with borrowed or shared words, such as proper nouns in Spanish (ES) and English (EN), pronunciations of the same text might be different.
This adds more %
ambiguity when a naively trained model sometimes generates accented %
speech for a particular speaker.

Zen et al. proposed a speaker and language factorization for HMM-based parametric TTS system \cite{zen2012slf}, aiming to transfer a voice from one language to others.
\cite{li2016multi} proposed a multilingual parametric neural TTS system, which used a unified input representation and shared parameters across languages, however the voices used for each language were disjoint. %
\cite{ming2017light} described a similar bilingual Chinese and English neural TTS system trained on speech from a bilingual speaker, allowing it to synthesize speech in both languages using the same voice.  \cite{lee2018learning} studied learning pronunciation from a bilingual TTS model. %
Most recently, \cite{nachmani2019unsupervised} presented a multilingual neural TTS model %
which supports voice cloning across English, Spanish, and German. 
It used language-specific text and speaker encoders, and incorporated a secondary fine-tuning step to optimize a speaker identity-preserving loss, ensuring that the model could output a consistent voice regardless of language.
We also note that the sound quality is not on par with recent neural TTS systems, potentially because of its use of the WORLD vocoder \cite{morise2016world} for waveform synthesis. %

Our work is most similar to \cite{li2018bytes}, which describes a multilingual TTS model based on Tacotron~2\cite{shen2018natural} which uses a Unicode encoding ``byte'' input representation to train a model on one speaker of each of English, Spanish, and Mandarin.
In this paper, we evaluate different input representations, scale up the number of training speakers for each language, and extend the model to support cross-lingual voice cloning.
The model is trained in a single stage, with no language-specific components,
and obtains naturalness on par with baseline monolingual models. %
Our contributions include: (1) Evaluating the effect of using different text input representations in a multilingual TTS model. (2) Introducing a per-input token speaker-adversarial loss to enable cross-lingual voice transfer when only one training speaker is available for each language. (3) Incorporating an explicit language embedding to the input, which enables moderate control of speech accent, independent of speaker identity, when the training data contains multiple speakers per language.

We evaluate the contribution of each component, and demonstrate the proposed model's ability to disentangle speakers from languages and consistently synthesize high quality speech for all speakers, despite the perfect correlation to the original language in the training data.

\section{Model Structure}
\label{sec:tts}
\begin{figure}[tb]
 \centering
 \centerline{\includegraphics[width=0.9\linewidth]{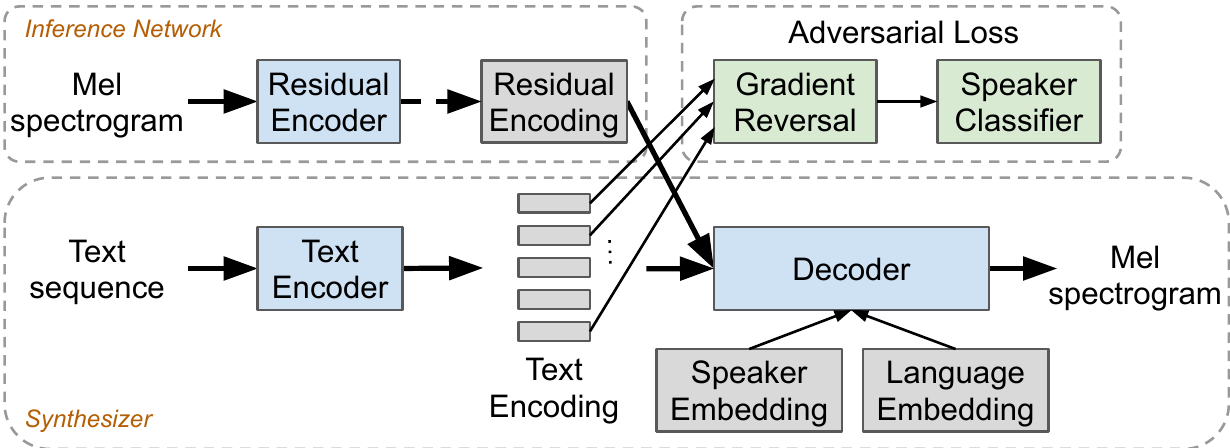}}
 \vskip-1ex
  \caption{Overview of the components of the proposed model. Dashed lines denote sampling via reparameterization~\cite{kingma2014auto} during training. The prior mean is always use during inference.}
  \label{fig:diagram}
  \vspace{-1.5ex}
\end{figure}

We base our multilingual TTS model on Tacotron~2~\cite{shen2018natural}, which uses an attention-based sequence-to-sequence model to generate a sequence of log-mel spectrogram frames based on an input text sequence.
The architecture is illustrated in Figure~\ref{fig:diagram}.
It augments the base Tacotron~2 model with additional speaker and, optionally, language embedding inputs (bottom right), an adversarially-trained speaker classifier (top right), and a variational autoencoder-style residual encoder (top left) which conditions the decoder on a latent embedding computed from the target spectrogram during training (top left).
Finally, similar to Tacotron~2, we separately train a WaveRNN~\cite{kalchbrenner2018efficient} neural vocoder.

\subsection{Input representations}
End-to-end TTS models have typically used character \cite{wang2017tacotron} or phoneme \cite{sotelo2017char2wav,wang2018style} input representations, or hybrids between them \cite{ping2018deep,kastner2019mix}. Recently, \cite{li2018bytes} proposed using inputs derived from the UTF-8 byte encoding in multilingual settings.
We evaluate the effects of using these representations for multilingual TTS.

\subsubsection{Characters / Graphemes} %
Embeddings corresponding to each character or grapheme are the default inputs for end-to-end TTS models~\cite{sotelo2017char2wav,wang2017tacotron,shen2018natural}, requiring the model to implicitly learn how to pronounce input words (i.e.\ grapheme-to-phoneme conversion \cite{van1993data}) as part of the synthesis task.
Extending a grapheme-based input vocabulary to a multilingual setting is straightforward, by simply concatenating grapheme sets in the training corpus for each %
language.
This can grow quickly for languages with large alphabets, %
e.g.\ our Mandarin vocabulary contains over 4.5k tokens. 
We simply concatenate all graphemes appearing in the training corpus, leading to a total of 4,619 tokens. Equivalent graphemes are shared across languages.
During inference all previously unseen characters are mapped to a special out-of-vocabulary (OOV) symbol.

\subsubsection{UTF-8 Encoded Bytes} %
Following \cite{li2018bytes} we experiment with an input representation based on the UTF-8 text encoding, which uses 256 possible values as each input token where the mapping from graphemes to bytes is language-dependent.
For languages with single-byte characters (e.g., English), this representation is equivalent to the grapheme representation.
However, for languages with multi-byte characters (such as Mandarin) the %
TTS model must learn to attend to a consistent sequence of bytes to correctly generate the corresponding speech.
On the other hand, using a UTF-8 byte representation may promote sharing of representations between languages due to the smaller number of input tokens.

\subsubsection{Phonemes} %
Using phoneme inputs simplifies the TTS task, as the model no longer needs to learn complicated pronunciation rules for languages such as English.
Similar to our grapheme-based model, equivalent phonemes are shared across languages.
We concatenate all possible phoneme symbols, for a total of 88 tokens.

To support Mandarin, we include tone information %
by learning phoneme-independent embeddings for each of the 4 possible tones, and broadcast each tone embedding to all phoneme embeddings inside the corresponding syllable. %
For English and Spanish, tone embeddings are replaced by stress embeddings which include primary and secondary stresses. A special symbol is used when there is no tone or stress.

\subsection{Residual encoder} %
Following \cite{hsu2018hierarchical}, we augment the TTS model by incorporating a variational autoencoder-like \emph{residual encoder} which encodes the latent factors in the training audio, e.g.\ prosody or background noise, which is not well-explained by the conditioning inputs: the text representation, speaker, and language embeddings.
We follow the structure from \cite{hsu2018hierarchical}, except we use a standard single Gaussian prior distribution and reduce the latent dimension to $16$.
In our experiments, we observe that feeding in the prior mean (all zeros) during inference, significantly improves stability of cross-lingual speaker transfer and leads to improved naturalness as shown by MOS evaluations in Section~\ref{sec:vae}.

\subsection{Adversarial training}
One of the challenges for multilingual TTS is data sparsity, where some languages may only have training data for a few speakers. In the extreme case where there is only one speaker per language in the training data, the speaker identity is essentially the same as the language ID.  
To encourage the model to learn disentangled representations of the text and speaker identity, we proactively discourage the text encoding $\bf{t}_s$ from also capturing speaker information.
We employ domain adversarial training~\cite{ganin2016domain} to encourage $\bf{t}_i$ to encode text in a speaker-independent manner by introducing a speaker classifier based on the text encoding and a gradient reversal layer.
Note that the speaker classifier is optimized with a different objective than the rest of the model: $\L_{\mathrm{speaker}}(\psi_{s}; \bf{t}_i) = \sum_{i}^N \log p(\bf{s}_i \mid \bf{t}_i)$, where $\bf{s}_i$ is the speaker label and $\psi_{s}$ are the parameters for speaker classifier. 
To train the full model, we insert a gradient reversal layer~\cite{ganin2016domain} prior to this speaker classifier, which scales the gradient by $-\lambda$. Following \cite{hsu2018disentangling}, we also explore inserting another adversarial layer on top of the variational autoencoder to encourage it to learn speaker-independent representations. However, we found that this layer has no effect after decreasing the latent space dimension.

We impose this adversarial loss separately on each element of the encoded text sequence, in order to encourage the model to learn a speaker- and language-independent text embedding space.
In contrast to \cite{hsu2018disentangling}, which disentangled speaker identity from background noise, some input tokens are highly language-dependent which can lead to unstable adversarial classifier gradients.  We address this by  clipping gradients computed at the reversal layer to limit the impact of such outliers.

\section{Experiments}%
\label{sec:results}

We train models using a proprietary dataset composed of high quality speech in three 
languages: \begin{inparaenum}[(1)] \item 385 hours of English (EN)
from 84 professional voice actors with accents from %
the United States, Great Britain, Australia, and Singapore; \item 97 hours of Spanish (ES) from 3 female speakers include Castilian and US Spanish; \item 68 hours of Mandarin (CN) from 5 speakers.\end{inparaenum}

\subsection{Model and training setup}
The synthesizer network  uses the Tacotron~2 architecture~\cite{shen2018natural}, with additional inputs consisting of learned speaker (64-dim) and language embeddings (3-dim), concatenated and passed to the decoder at each step.
The generated speech is represented as a sequence of 128-dim log-mel spectrogram frames, computed from 50ms windows shifted by 12.5ms.

The variational residual encoder architecture closely follows the attribute encoder in~\cite{hsu2018hierarchical}.
It maps a variable length mel spectrogram to two vectors parameterizing the mean and log variance of the Gaussian posterior. %
The speaker classifiers are fully-connected networks with one 256 unit hidden layer followed by a softmax predicting the speaker identity. The synthesizer and speaker classifier are trained with weight $1.0$ and $0.02$ respectively.
As described in the previous section we apply gradient clipping with factor $0.5$ to the gradient reversal layer. %

The entire model is trained jointly with a batch size of 256, using the Adam optimizer configured with an initial learning rate of $10^{-3}$, and an exponential decay that halves the learning rate every 12.5k steps, starting at 50k steps. 

Waveforms are synthesized using a WaveRNN~\cite{kalchbrenner2018efficient} vocoder which generates 16-bit signals sampled at 24~kHz conditioned on spectrograms predicted by the TTS model. We 
synthesize 100 samples per model, and have each one rated by 6 raters.

\subsection{Evaluation}
\begin{table}[t]
\caption{Speaker similarity Mean Opinion Score (MOS) comparing ground truth audio from speakers of different languages.  %
Raters are native speakers of the target language.}
\vskip-0.5ex
\centering
\begin{tabular}{lccc}
\toprule
\multirow[b]{2}{4em}[-1.15ex]{\tbh{Source Language}} & \multicolumn{3}{c}{\tbh{Target Language}} \\
\cmidrule(lr){2-4}
& \tbh{EN} & \tbh{ES} & \tbh{CN}\\
\midrule
EN & 4.40$\pm$0.07 & \dg{1.72$\pm$0.15} & \dg{1.80$\pm$0.08}  \\ %
ES & \dg{1.49$\pm$0.06} & 4.39$\pm$0.06 & \dg{2.14$\pm$0.09} \\
CN & \dg{1.32$\pm$0.06} & \dg{2.06$\pm$0.09} & 3.51$\pm$0.12\\
\bottomrule
\end{tabular}
\vskip-1.5ex
\label{tbl:spk_gt-sim}
\end{table}

To evaluate synthesized speech, %
we rely
on crowdsourced Mean Opinion Score (MOS) evaluations of \emph{speech naturalness} via %
subjective listening tests. Ratings follow the %
Absolute Category Rating scale, with scores from 1 to 5
in 0.5 point increments.

For cross-language voice cloning, %
we also evaluate whether the synthesized speech resembles the identity of the reference speaker by pairing each synthesized utterance with a reference utterance from the same speaker for subjective MOS evaluation of \emph{speaker similarity}, as in \cite{jia2018transfer}.
Although rater instructions explicitly asked for the content to be ignored,
note that this similarity evaluation is more challenging than the one in \cite{jia2018transfer} because the reference and target examples are spoken in different languages, and raters are not bilingual. %
We found that low fidelity audio tended to result in high variance similarity MOS so we always use WaveRNN outputs.\footnote{Some raters gave low fidelity audio lower scores, treating "blurriness" as a property of the speaker.  Others gave higher scores because they recognized such audio as synthetic and had lower expectations.}

For each language, we chose one speaker to use for similarity tests. As shown in Table~\ref{tbl:spk_gt-sim}, the EN speaker is found to be dissimilar to the ES and CN speakers (MOS below 2.0), while the ES and CN speakers are slightly similar (MOS around 2.0). The CN speaker has more natural variability compared to EN and ES, leading to a lower self similarity.
The scores are consistent when EN and CN raters evaluate the same EN and CN test set. The observation is consistent with \cite{wester2011cross}: raters are able to discriminate between speakers across languages. However, when rating synthetic speech, we observed that English speaking raters often considered ``heavy accented'' synthetic CN speech to sound more similar to the target EN speaker, compared to more fluent speech from the same speaker. This indicates that accent and speaker identity are not fully disentangled. %
We encourage readers to listen to samples on the companion webpage.\footnote{\smallurl{
http://google.github.io/tacotron/publications/multilingual}}%

\subsection{Comparing input representations}%
\begin{table}[!t]
\caption{Naturalness MOS of monolingual and multilingual models synthesizing speech of in different languages.} %
\vskip-0.5ex
\centering
\setlength{\tabcolsep}{1.1ex}
\begin{tabular}{l@{\hspace{1ex}}l@{\hspace{1ex}}ccc}
\toprule
& & \multicolumn{3}{c}{\tbh{Language}} \\
\cmidrule(lr){3-5}
\tbh{Model} & \tbh{Input} & \tbh{EN} & \tbh{ES} & \tbh{CN} \\
\midrule
Ground truth &         & 4.60$\pm$0.05 &   4.37$\pm$0.06   & 4.42$\pm$0.06\\
\midrule
Monolingual &
char & 4.24$\pm$0.12  & 4.21$\pm$0.11 & 3.48$\pm$0.11 \\
& phone & 4.59$\pm$0.06 & {4.39$\pm$0.04} & 4.16$\pm$0.08\\
\addlinespace %
Multilingual &
byte & 4.23$\pm$0.14  & 4.23$\pm$0.10 & 3.42$\pm$0.12 \\ %
1EN 1ES 1CN & char & 3.94$\pm$0.15 & 4.33$\pm$0.09 & 3.63$\pm$0.10 \\
& phone & 4.34$\pm$0.09 & 4.41$\pm$0.05 & 4.06$\pm$0.10\\
\addlinespace
Multilingual & byte & 4.11$\pm$0.14 & 4.21$\pm$0.12 &  3.67$\pm$0.12\\
84EN 3ES 5CN & char & 4.26$\pm$0.13  & 4.23$\pm$0.11 & 3.46$\pm$0.11\\
& phone & 4.37$\pm$0.12 & 4.37$\pm$0.04 & 4.09$\pm$0.10\\
\bottomrule
\end{tabular}
\vskip-0.25ex
\label{tbl:multitts-mos}
\end{table}
\begin{table}[!t]
	\centering
    \caption{  Naturalness and speaker similarity MOS of cross-language voice cloning of an EN source speaker.  Models which use different input representations are compared, with and without the speaker-adversarial loss. fail: raters complained that too many utterances were spoken in the wrong %
    language.}
    \vskip-0.5ex
	\label{tab:transfer1}
    \setlength{\tabcolsep}{0.4em}
    \begin{tabular}{lcc cc}
		\toprule
	     & \multicolumn{2}{c}{ES target} & \multicolumn{2}{c}{CN target} \\
        \cmidrule(lr){2-3} \cmidrule(lr){4-5}
		\hspace{0.5em}Input & Naturalness & Similarity & Naturalness & Similarity \\
		\midrule
        \hspace{0.5em}char & 2.62$\pm$0.10 &4.25$\pm$0.09 & N/A & N/A  \\
        \hspace{0.5em}byte & 2.62$\pm$0.15 & \href{https://furball.corp.google.com/project/load?projectId=122532779&tab=project}{3.96$\pm$0.10}  & N/A & N/A  \\
    	\addlinespace
    	\multicolumn{5}{l}{with adversarial loss} \\
    	\hspace{0.5em}byte & 2.34$\pm$0.10 & 4.23$\pm$0.09 & fail & 3.85$\pm$0.11 \\ %
        \hspace{0.5em}phone & 3.20$\pm$0.09\cmmnt{2.54$\pm$0.08} & 4.15$\pm$0.10 & 2.75$\pm$0.12 & 3.60$\pm$0.09\\
        \bottomrule
	\end{tabular}%
	\vskip-1.5ex
\end{table}
\begin{table*}[t]
	\centering
	\caption{Naturalness and speaker similarity MOS of cross-language voice cloning of the full multilingual model using phoneme inputs.}
	\label{tab:alltransfer}
    \vskip-0.5ex
    \setlength{\tabcolsep}{0.5em}
    \begin{tabular}{llcc cc cc}
		\toprule
		\multirow[b]{2}{4em}[-1.15ex]{Source Language}
		&  & \multicolumn{2}{c}{EN target}   & \multicolumn{2}{c}{ES target} & \multicolumn{2}{c}{CN target} \\
        \cmidrule(lr){3-4} \cmidrule(lr){5-6} \cmidrule(lr){7-8}
		& Model & Naturalness & Similarity & Naturalness & Similarity & Naturalness & Similarity \\
        \midrule
        - & Ground truth (self-similarity) & 4.60$\pm$0.05 & 4.40$\pm$0.07
                         & 4.37$\pm$0.06 & 4.39$\pm$0.06
                         & 4.42$\pm$0.06 & 3.51$\pm$0.12 \\
		\midrule
		               {EN} 
        		            & 84EN 3ES 5CN & 4.37$\pm$0.12 & 4.63$\pm$0.06 & 4.20$\pm$0.07 & 3.50$\pm$0.12 & 3.94$\pm$0.09 & 3.03$\pm$0.10\\
        		            & \hspace{1em}language ID fixed to EN & - & - & 3.68$\pm$0.07 & 4.06$\pm$0.09 & 3.09$\pm$0.09 & 3.20$\pm$0.09 \\
		\addlinespace
		               {ES} 
        		            & 84EN 3ES 5CN & 4.28$\pm$0.10 & 3.24$\pm$0.09 & 4.37$\pm$0.04 & 4.01$\pm$0.07 & 3.85$\pm$0.09 & 2.93$\pm$0.12\\

		\addlinespace
		               {CN} 
        		            & 84EN 3ES 5CN & 4.49$\pm$0.08 & 2.46$\pm$0.10 & 4.56$\pm$0.08 & 2.48$\pm$0.09 & 4.09$\pm$0.10 & 3.45$\pm$0.12 \\

        \bottomrule
	\end{tabular}%
	\vskip-1.5ex
\end{table*}
We first build and evaluate models comparing the performance of different text input representations. 
For all
three languages, byte-based models always use a 256-dim softmax output. Monolingual character and phoneme models each use a different input vocabulary corresponding to the training language.

Table~\ref{tbl:multitts-mos} compares monolingual and multilingual model performance using different input representations. For Mandarin, the phoneme-based model performs significantly better than char- or byte-based variants due to rare and OOV words. Compared to the monolingual system, multilingual phoneme-based systems have similar performance on ES and CN but are slightly worse on EN. 
CN has a larger gap to ground truth (top) due to unseen word segmentation (for simplicity, we didn't add word boundary during training). 
The multispeaker model (bottom) performs about the same as the single speaker per-language variant (middle).
Overall, when using phoneme inputs all the languages obtain MOS scores above 4.0.

\subsection{Cross-language voice cloning} %

We evaluate how well the multispeaker models can be used to clone a speaker's voice into a new language by simply passing in speaker embeddings corresponding to a different language from the input text.
Table~\ref{tab:transfer1} shows voice cloning performance from an EN speaker in the most data-poor scenario (129 hours), where only a single speaker is available for each training language (1EN 1ES 1CN) without using the speaker-adversarial loss. 
Using byte inputs \footnote{Using character or byte inputs led to similar results.} it was possible to clone the EN speaker to ES with high similarity MOS, albeit with significantly reduced naturalness.
However, cloning the EN voice to CN failed\footnote{We didn't run listening tests because it was clear that synthesizing EN text using the CN speaker embedding didn't affect the model output.}, as did 
cloning to ES and CN using phoneme inputs. %

Adding the adversarial speaker classifier enabled cross-language cloning of the EN speaker to CN with very high similarity MOS for both  byte and phoneme models.
However, naturalness MOS remains much lower than using the native speaker identity,  with the naturalness listening test failing entirely in the CN case with byte inputs as a result of rater comments that the speech sounded like a foreign language. 
According to rater comments on the phoneme system, most of the degradation came from mismatched accent and pronunciation, not fidelity. %
CN raters commented that it sounded like ``a foreigner speaking Chinese''. More interestingly, few ES raters commented that ``The voice does not sound robotic but instead sounds like an English native speaker who is learning to pronounce the words in Spanish.''
Based on these results, we only use phoneme inputs in the following experiments since this
guarantees that pronunciations are correct and results in more fluent speech.

Table~\ref{tab:alltransfer} evaluates %
voice cloning performance of the full multilingual model (84EN 3ES 5CN), which is
 trained on the full dataset with increased speaker coverage, %
and uses the speaker-adversarial loss and speaker/language embeddings.
Incorporating the adversarial loss forces the text representation to be less language-specific, instead relying on the language embedding to capture language-dependent information.
Across all language pairs, the model synthesizes speech in all voices with naturalness MOS above 3.85, demonstrating that increasing training speaker diversity improves generalization.
In most cases synthesizing EN and ES speech (except EN-to-ES) approaches the ground truth scores. %
In contrast, naturalness of CN speech is consistently lower than the ground truth.

The high naturalness and similarity MOS scores in the top row of Table~\ref{tab:alltransfer} indicate that the model is able to successfully transfer the EN voice to both ES and CN %
almost without  accent.
When consistently conditioning on the EN language embedding regardless of the target language (second row), the model produces more English accented ES and CN speech, which leads to lower naturalness but higher similarity MOS scores.
Also see Figure \ref{fig:dvector-visualization}
and the demo for accent transfer audio examples.

We see that cloning the CN voice to other languages (bottom row) has the lowest similarity MOS, although the scores are still much higher than different-speaker similarity MOS in the off-diagonals of Table~\ref{tbl:spk_gt-sim} indicating that there is some degree of transfer.
This is a consequence of the low speaker coverage of CN compared to EN in the training data, as well as the large distance between CN and other languages.

\label{sec:vae}
\begin{table}[t]
\caption{Effect of EN speaker cloning with no residual encoder.} %
\vskip-0.5ex
\centering
\setlength{\tabcolsep}{0.5em}
\begin{tabular}{lccc}
\toprule
& \multicolumn{3}{c}{\tbh{Target Language}} \\
\cmidrule(lr){2-4}
\tbh{Model} & \tbh{EN} & \tbh{ES} & \tbh{CN} \\
\midrule
84EN 3ES 5CN & 4.37$\pm$0.12 & 4.20$\pm$0.07 & 3.94$\pm$0.09\\
\hspace{1em}- residual encoder & 4.38$\pm$0.10 & 4.11$\pm$0.06 &  3.52$\pm$0.11 \\ %
\bottomrule
\end{tabular}
\label{tbl:vae-mos}
\end{table}
Finally, Table~\ref{tbl:vae-mos} demonstrates the importance of training using a variational residual encoder to stabilize the model output.  Naturalness MOS decreases by 0.4 points for EN-to-CN cloning without the residual encoder (bottom row).
In informal comparisons of the outputs of the two models we find that the model without the residual encoder tends to skip rare words or inserts unnatural pauses in the output speech. This indicates the VAE prior learns a mode which helps stabilize attention. %

\iftrue
\begin{figure}[t]
 \centering
 \vskip-1ex
 \centerline{\includegraphics[width=0.93\linewidth]{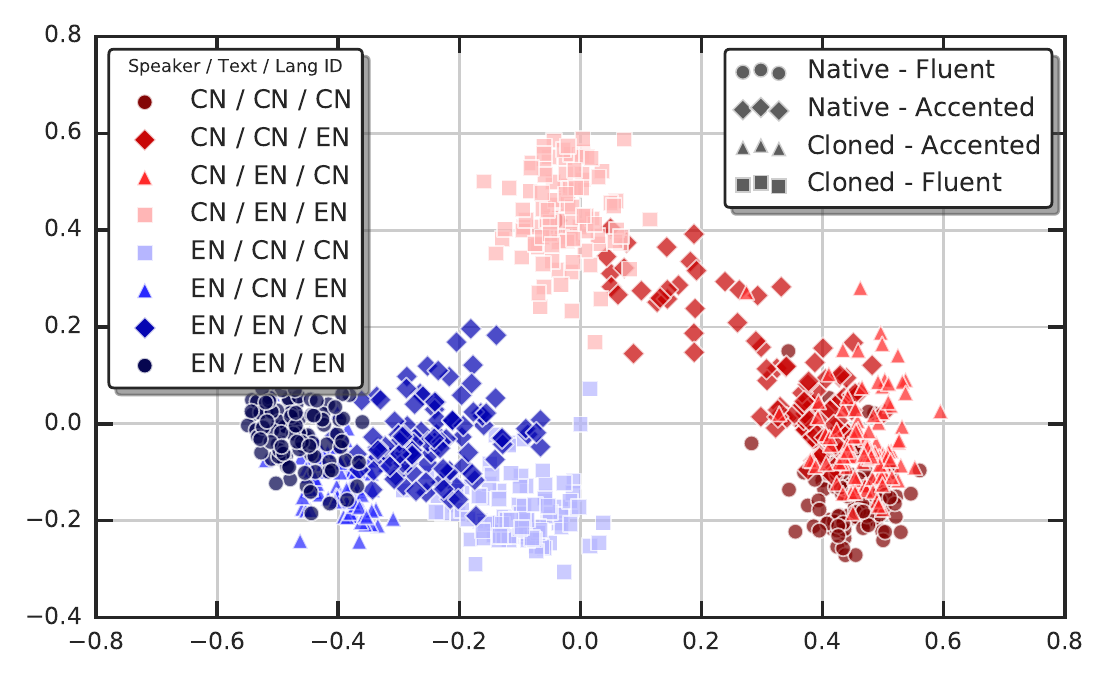}}
 \vskip-2ex
  \caption{
    Visualizing the effect of voice cloning and accent control, using 2D PCA of speaker embeddings \cite{wan2018generalized} computed from speech synthesized %
    with different speaker, text, and language ID combinations. %
    Embeddings cluster together (bottom left and right), implying high similarity, when the speaker's original language matches the language embedding, regardless of the text language.
    However, using language ID from the text (squares), modifying the speaker's accent to speak fluently, hurts similarity compared to the native language and accent (circles).
  }
  \label{fig:dvector-visualization}
  \vskip-3ex
\end{figure}
\fi

\section{Conclusions}
\label{sec:concl}
We describe extensions to the Tacotron~2 neural TTS model which allow training of a  multilingual model trained only on monolingual speakers, which is able to synthesize high quality speech in three languages, and transfer training voices across languages.
Furthermore, the model learns to speak foreign languages with moderate control of accent, and, as demonstrated on the companion webpage, has rudimentary support for code switching.
In future work we plan to investigate methods for scaling up to leverage large amounts of low quality training data, and support many more speakers and languages.

\section{Acknowledgements}
We thank Ami Patel, Amanda Ritchart-Scott, Ryan Li, Siamak Tazari, Yutian Chen, Paul McCartney, Eric Battenberg, Toby Hawker, and Rob Clark for discussions and helpful feedback.

\vfill\pagebreak

\bibliographystyle{IEEEtran}
\bibliography{refs}

\end{document}